\documentclass[]{article}
\usepackage[letterpaper]{geometry}
\usepackage{mtsummit2015}
\usepackage{times}
\usepackage{url}
\usepackage{latexsym}
\usepackage{natbib}
\usepackage{layout}
\usepackage{graphicx}
\usepackage{color}
\usepackage{booktabs}
\usepackage{caption}
\usepackage{subfigure}
\usepackage{multirow}
\usepackage[textfont=it]{caption}
\usepackage{comment}
\usepackage{float}

\begin{document}

\title{\bf OpenNMT: Neural Machine Translation Toolkit}

\author{\name{\bf Guillaume Klein} \hfill \addr{SYSTRAN}
\AND
        \name{\bf Yoon Kim} \hfill \addr{Harvard University}
\AND
       \name{\bf Yuntian Deng} \hfill \addr{Harvard University}
        \AND
       \name{\bf Vincent Nguyen} \hfill \addr{Ubiqus}
        \AND
       \name{\bf Jean Senellart} \hfill \addr{SYSTRAN}
        \AND
       \name{\bf Alexander M. Rush} \hfill \addr{Harvard University}
}

\maketitle
\pagestyle{empty}

\begin{abstract}
  OpenNMT is an open-source toolkit for neural machine translation
  (NMT).  The system prioritizes efficiency, modularity, and
  extensibility with the goal of supporting NMT research into model
  architectures, feature representations, and source modalities, while
  maintaining competitive performance and reasonable training
  requirements. The toolkit consists of modeling and translation
  support, as well as detailed pedagogical documentation about the
  underlying techniques. OpenNMT has been used in several production
  MT systems, modified for numerous research papers, and is
  implemented across several deep learning frameworks.
\end{abstract}

\section{Introduction}

Neural machine translation (NMT) is a new methodology for machine
translation that has led to remarkable improvements, particularly in
terms of human evaluation, compared to rule-based
and statistical machine translation (SMT) systems
\citep{wu2016google,systran}. Originally developed using pure
sequence-to-sequence models \citep{sutskever14sequence,Cho2014} and
improved upon using attention-based variants \citep{Bahdanau2015,Luong2015}, NMT has now become a widely-applied technique for machine
translation, as well as an effective approach for other related NLP
tasks such as dialogue, parsing, and summarization.

As NMT approaches are standardized, it becomes more important for the
machine translation and NLP community to develop open implementations
for researchers to benchmark against, learn from, and extend
upon. Just as the SMT community benefited greatly from toolkits like
Moses \citep{koehn2007moses} for phrase-based SMT and CDec
\citep{dyer2010cdec} for syntax-based SMT, NMT toolkits can provide a
foundation to build upon. A toolkit should aim to provide
a shared framework for developing and comparing open-source systems,
while at the same time being efficient and accurate enough to be used
in production contexts.

\begin{figure*}[t]
  \centering
  \subfigure[]{%
    \includegraphics[width=0.5\textwidth]{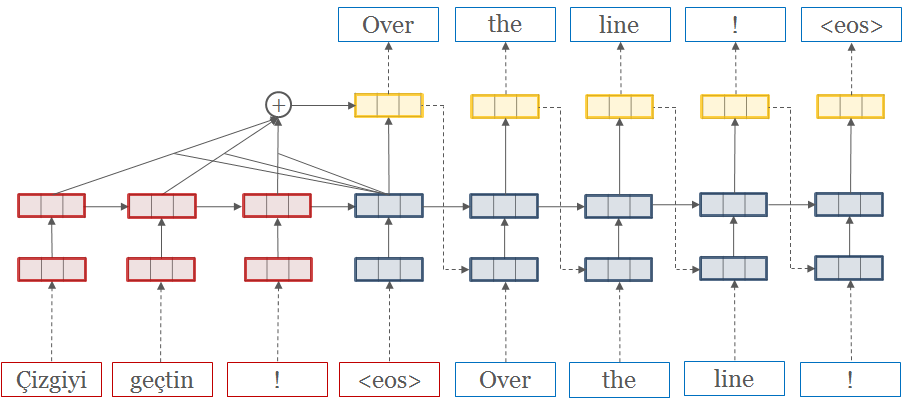}%
    \label{fig:rnn}%
    }
    \subfigure[]{%
    \includegraphics[width=0.5\textwidth]{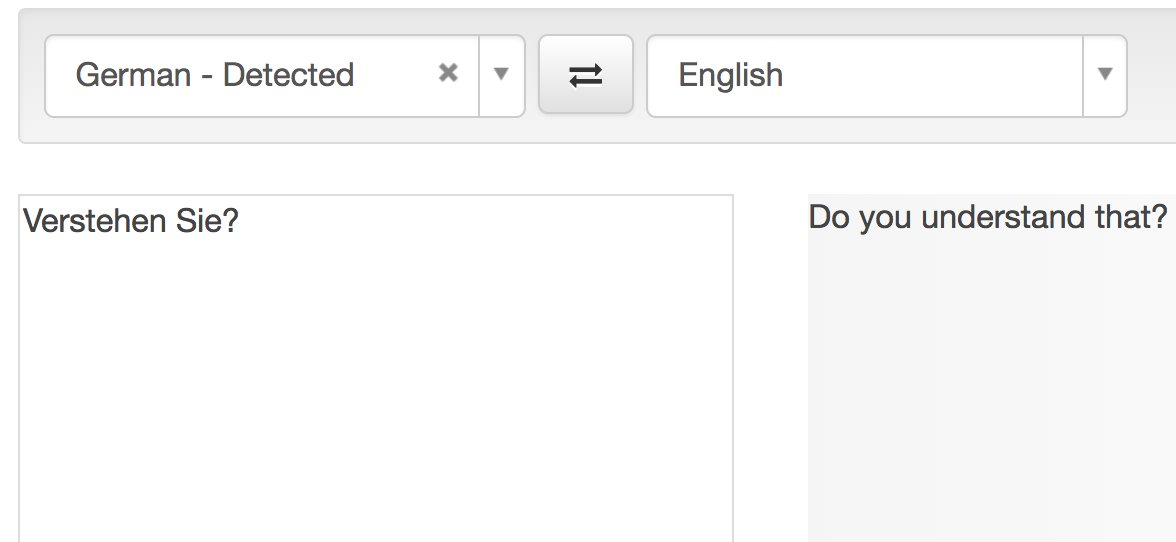}%
    \label{fig:live}%
  }%
  \vspace{-0.3cm}
  \caption{\small (a). Schematic view of neural machine translation. The \textcolor{red}{red} source words are first mapped to word vectors and then fed into a recurrent neural network (RNN). Upon seeing the $\langle$eos$\rangle$ symbol, the final time step initializes a target \textcolor{blue}{blue} RNN. At each target time step, \textit{attention} is applied over the source RNN and combined with the current hidden state to produce a prediction $p(w_t| w_{1: t-1}, x)$ of the next word. This prediction is then fed back into the target RNN. (b). Live demo of the OpenNMT system.}
  \label{fig:ab}
\end{figure*}

With these goals in mind, in this work we present an open-source toolkit for developing
neural machine translation systems, known as \textit{OpenNMT} (\url{http://opennmt.net}). Since its launch in December 2016, OpenNMT has become a collection of implementations targeting both academia and industry. The system is designed to be simple to use and easy to extend, while maintaining efficiency and state-of-the-art accuracy. In
addition to providing code for the core translation tasks, OpenNMT was
designed with two aims: (a) prioritize training and test
efficiency, (b) maintain model modularity and readability hence research extensibility.

During this time, many other stellar open-source NMT implementations
have also been released, including \textit{GroundHog},
\textit{Blocks}, \textit{Nematus}, \textit{tensorflow-seq2seq},
\textit{GNMT}, \textit{fair-seq},  \textit{Tensor2Tensor}, \textit{Sockeye},
\textit{Neural Monkey}, \textit{lamtram}, \textit{XNMT},
\textit{SGNMT}, and \textit{Marian}. These projects mostly implement
variants of the same underlying systems, and differ in their
prioritization of features. The open-source community around
this area is flourishing, and is providing the NLP community a useful variety of open-source NMT frameworks.
In the ongoing development of OpenNMT, we aim to build upon the
strengths of those systems, while supporting a framework with
high-accuracy translation, multiple options and clear documentation.

This engineering report describes how the system targets our design
goals. We begin by briefly surveying the background for NMT, and then
describing the high-level implementation details. We end by showing
benchmarks of the system in terms of accuracy, speed, and memory usage
for several translation and natural language generation tasks.

\section{Background}

NMT has now been extensively described in many
excellent tutorials (see for instance
\url{https://sites.google.com/site/acl16nmt/home}). We give only
a condensed overview.

NMT takes a conditional language modeling view of translation by modeling the
probability of a target sentence $w_{1:T}$ given a source sentence
$x_{1:S}$ as
$p(w_{1:T}| x) = \prod_{1}^T p(w_t| w_{1:t-1}, x; \theta)$ where the 
distribution is parameterized with $\theta$.
This
distribution is estimated using an attention-based encoder-decoder
architecture \citep{Bahdanau2015}. A source encoder recurrent neural
network (RNN) maps each source word to a word vector, and processes
these to a sequence of hidden vectors
$\mathbf{h}_1, \ldots, \mathbf{h}_S$.  The target decoder combines an
RNN hidden representation of previously generated words
($w_1, ... w_{t-1}$) with source hidden vectors to predict scores for
each possible next word. A softmax layer is then used to produce a
next-word distribution $ p(w_t| w_{1:t-1}, x; \theta)$. The source
hidden vectors influence the distribution through an attention pooling
layer that weights each source word relative to its expected
contribution to the target prediction. The complete model is trained
end-to-end to minimize the negative log-likelihood of the training
corpus. An unfolded network diagram is shown in Figure~\ref{fig:rnn}.

In practice, there are also many other important aspects that improve
the effectiveness of the base model. Here we briefly mention four
areas: (a) It is important to use a gated RNN such as an LSTM
\citep{hochreiter1997long} or GRU \citep{chung2014empirical} which help
the model learn long-term features. (b) Translation requires
relatively large, stacked RNNs, which consist of several vertical
layers (2-16) of RNNs at each time step \citep{sutskever14sequence}. (c)
Input feeding, where the previous attention vector is fed back into
the input as well as the predicted word, has been shown to be quite
helpful for machine translation \citep{Luong2015}.  (d) Test-time
decoding is done through \textit{beam search} where multiple
hypothesis target predictions are considered at each time
step. Implementing these correctly can be difficult, which motivates
their inclusion in a NMT framework.

\section{Implementation}

OpenNMT is a community of projects supporting easy adoption neural
machine translation. At the heart of the project are libraries for
training, using, and deploying neural machine translation models. The
system was based originally on \textit{seq2seq-attn}, which was rewritten for ease of efficiency, readability, and
generalizability. The project supports vanilla NMT models along with
support for attention, gating, stacking, input feeding,
regularization, copy models, beam search and all other options
necessary for state-of-the-art performance.

OpenNMT has currently three main implementations. All of them are actively maintained:

\begin{itemize}
\item \textit{OpenNMT-lua} The original project developed in Torch 7.
  Full-featured, optimized, and stable code ready for quick
  experiments and production.
  \item \textit{OpenNMT-py} An OpenNMT-lua clone using PyTorch.
    Initially created by by Adam Lerer and the Facebook AI research
    team as an example, this implementation is easy to extend and
    particularly suited for research.
  \item \textit{OpenNMT-tf} An implementation following the style of
    TensorFlow.  This is a newer project focusing on large scale
    experiments and high performance model serving using the latest
    TensorFlow features.
\end{itemize}

OpenNMT is developed completely in the open on GitHub at
(\url{http://github.com/opennmt}) and is MIT licensed.  The initial
release has primarily contributions from SYSTRAN Paris, the Harvard
NLP group and Facebook AI research. Since official beta release, the
project (OpenNMT-lua, OpenNMT-py and OpenNMT-tf) has been starred by
over 2500 users in total, and there have been over 100 outside
contributors. The project has an active forum for community feedback
with over five hundred posts in the last two months. There is also a
live demonstration available of the system in use
(Figure~\ref{fig:live}).

One often overlooked benefit of  NMT compared to SMT is its relative compactness.
OpenNMT-lua including preprocessing and model variants is roughly 16K lines of code,
the PyTorch version is less than 4K lines and Tensorflow version has around 7K lines. For comparison the Moses SMT framework
including language modeling is over 100K lines. This makes our system
easy to completely understand for newcomers. Each project is fully
self-contained depending on minimal number of external libraries
and also includes some preprocessing, visualization and analysis tools.

\section{Design Goals}
\begin{figure}[t]
        \centering
        \includegraphics[scale=0.3]{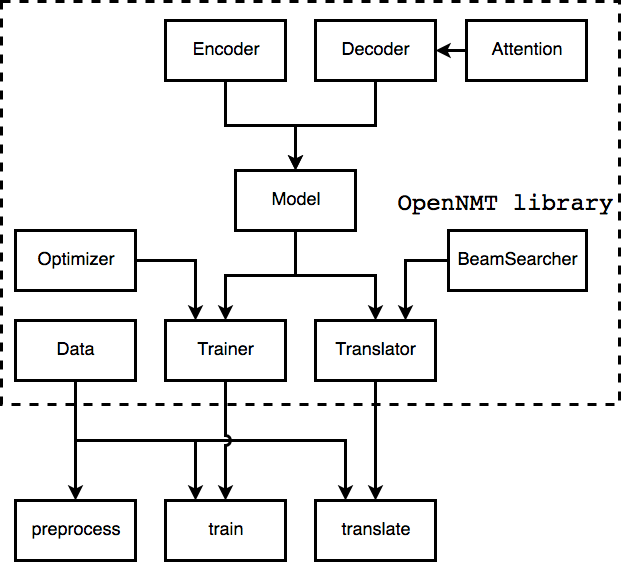}
        \caption{\label{fig:codestruct}Schematic overview of OpenNMT-py code}
    \end{figure}
    
\subsection{System Efficiency}

As NMT systems can take from days to weeks to train, training
efficiency is a paramount concern. Slightly faster training can make the difference between
plausible and impossible experiments.

\paragraph{Memory Sharing \& Sharding}

When training GPU-based NMT models, memory size restrictions are the
most common limiter of batch size, and thus directly impact training
time. Neural network toolkits, such as Torch, are often designed to
trade-off extra memory allocations for speed and declarative
simplicity. For OpenNMT, we wanted to have it both ways, and so we
implemented an external memory sharing system that exploits the known
time-series control flow of NMT systems and aggressively shares the
internal buffers between clones. The potential shared buffers are
dynamically calculated by exploration of the network graph before
starting training. In practical use, aggressive memory reuse provides
a saving of 70\% of GPU memory with the default model size. For OpenNMT-py, we implemented a sharding mechanism both for data loading to enable training on extremely large datasets that cannot fit into memory, and for back-propagation to reduce memory footprints during training.

\paragraph{Multi-GPU} OpenNMT additionally supports multi-GPU training
using data parallelism. Each GPU has a replica of the master
parameters and processes independent batches during training phase.  Two
modes are available: synchronous and asynchronous training \citep{dean2012large}. Experiments with 8 GPUs show a 6$\times$ speed up in
per epoch, but a slight loss in training efficiency. When training to similar
loss, it gives a 3.5$\times$ total speed-up to training.

\paragraph{C/Mobile/GPU Translation} Training NMT systems requires
significant code complexity to facilitate fast
back-propagation-through-time. At deployment, the system is much less
complex, and only requires (i) forwarding values through the network
and (ii) running a beam search that is much simplified compared to
SMT. OpenNMT includes several different translation deployments
specialized for different run-time environments: a batched CPU/GPU
implementation for very quickly translating a large set of sentences,
a simple single-instance implementation for use on mobile devices, and
a specialized C implementation suited for industrial use.

\subsection{Modularity for Research}

A secondary goal was a desire for code readability and extensibility.
We targeted this goal by explicitly separating training, optimization and different components of the model, and by including tutorial documentation within
the code. A schematic overview of our data structures in OpenNMT-py is shown in Figure~\ref{fig:codestruct}. We provide users with simple interfaces \textit{preprocess}, \textit{train} and \textit{translate}, which only require source/target files as input, while we provide a highly modularized library for advanced users.  Each module in the library is highly customizable and configurable with multiple ready-for-use features. Advanced users can access the modules directly through a library interface to construct and train variant of the standard NMT setup.

\begin{table}
\centering
          \begin{tabular}{l r}
          \toprule
            { System} & { BLEU-cased} \\
            \midrule
            uedin-nmt-ensemble & 28.3 \\
            LMU-nmt-reranked-wmt17-en-de & 27.1 \\
             SYSTRAN-single (OpenNMT) &  26.7 \\
            \bottomrule
          \end{tabular}
          \caption{\label{tab:wmt}Top 3 on English-German \emph{newstest2017} WMT17.}
\end{table}

\begin{table}
\parbox{.48\linewidth}{
\centering
\begin{tabular}{ccccc}
    \toprule
      System & \multicolumn{2}{c}{Speed tok/sec}  & BLEU\\
       & Train  & Trans  &  \\
    \midrule
  Nematus & 3221& 252 & 18.25 \\
    ONMT &5254 & 457 & 19.34\\
    \bottomrule
  \end{tabular}

  \caption{ \small \label{tab:res} Performance results for EN$\rightarrow$DE on WMT15 tested on \textit{newstest2014}. Both systems 2x500 RNN, embedding size 300, 13 epochs, batch size 64, beam size 5. We compare on a 32k BPE setting.}
}
\hfill
\parbox{.48\linewidth}{
\centering
\begin{tabular}{crrrrr}
    \toprule
     System & newstest14  & newstest17\\
    \midrule
     seq2seq & 22.19\ \  & - \\
    Sockeye & - & 25.55 \ \\
    ONMT & 23.23 [19.34] & 25.06 [22.69] &  \\
    \bottomrule
  \end{tabular}

  \caption{ \small \label{tab:results-updated} OpenNMT's performance as reported by \cite{britz2017massive} and \cite{hieber2017sockeye} (bracketed) compared to our best results. ONMT used 32k BPE, 2-layers bi-RNN of 1024, embedding size 512, dropout 0.1 and max length 100.}
}
\end{table}
\paragraph{Extensible Data, Models, and Search} In addition to plain text,
OpenNMT also supports different input types including models with
discrete features \citep{sennrich2016linguistic}, models with
non-sequential input such as tables, continuous data such as speech
signals, and multi-dimensional data such as images. To support these
different input modalities the library implements image encoder \citep{xu2015show,deng2017image} and
audio encoders \citep{DBLP:journals/corr/ChanJLV15}.  OpenNMT
implements various attention types including general, dot product, and
concatenation \citep{Luong2015,britz2017massive}.  This also
includes recent extensions to these standard modules such as the copy
mechanism \citep{vinyals2015pointer,gu2016incorporating}, which is
widely used in summarization and generation applications.

The newer implementations of OpenNMT have also  been updated to
include support for recent innovations in  non-recurrent translation models. In
particular recent support has been added for convolution translation
\citep{gehring2017convolutional} and the attention-only transformer
network \citep{vaswani2017attention}.

Finally, the translation code allows for user customization.
In addition to out-of-vocabulary
(OOV) handling \citep{luong2015b}, OpenNMT also allows beam search
with various normalizations including length and attention coverage
normalization \citep{wu2016google}, and dynamic dictionary support for
copy/pointer networks. We also provide an interface for customized
hypothesis filtering, enabling beam search under various constraints
such as maximum number of OOV's and lexical constraints.

\paragraph{Modularity}
Due to the deliberate modularity of our code, OpenNMT is readily extensible for novel feature development. As one example, by substituting the attention module, we can implement local attention \citep{Luong2015}, sparse-max attention
\citep{martins2016softmax} and structured attention \citep{kim2017structured} with minimal change of code. As another example, in order to get feature-based factored neural translation \citep{sennrich2016linguistic} we simply need to modify the input network to process the feature-based representation, and the output network to produce multiple conditionally independent predictions.

We have seen instances of this use in published research. In addition to
machine translation
\citep{levin2017toward,ha2017effective,ma2017osu},
researchers have employed OpenNMT for parsing \citep{van2017neural},
document summarization \citep{ling2017coarse}, data-to-document
\citep{wiseman2017challenges,gardent2017webnlg}, and transliteration
\citep{ameur2017arabic}, to name a few of many applications.

\paragraph{Additional Tools}

OpenNMT packages several additional tools, including: 1) reversible tokenizer, which can also perform Byte Pair Encoding (BPE) \citep{DBLP:journals/corr/SennrichHB15}; 2) loading and exporting word embeddings; 3) translation server which enables showcase results remotely; and 4) visualization tools for debugging or understanding, such as beam search visualization, profiler and TensorBoard logging.

\begin{table}
\parbox{.48\linewidth}{
\centering
\begin{tabular}{l cc}
          \toprule
            { System} & { newstest14}  & newstest15\\
            \midrule
            GNMT 4 layers & 23.7 & 26.5 \\
            GNMT 8 layers & 24.4 & 27.6 \\
            WMT reference & 20.6 & 24.9 \\
            ONMT & 23.2 & 26.0 \\
            \bottomrule
          \end{tabular}
          \caption{\small\label{tab:results-google}Comparison with \textit{GNMT} on EN$\rightarrow$DE. \textit{ONMT} used 2-layers bi-RNN of 1024, embedding size 512, dropout 0.1 and max length 100.}
         
}
\hfill
\parbox{.48\linewidth}{
\centering
\begin{tabular}{l cc}
          \toprule
            { System} & { newstest14}  & newstest17 \\
            \midrule
            T2T  & 27.3 & 27.8 \\
            ONMT T2T & 26.8 & 28.0 \\
            GNMT (rnn) & 24.6 & - \\
            ONMT (rnn) & 23.2 & 25.1 \\
            \bottomrule
          \end{tabular}
          \caption{\small\label{tab:results-transformer}Transformer Results on English-German newstest14 and newstest17. We use 6-layer transformer with model size of 512. }
        
}
\end{table}

\section{Experiments}

OpenNMT achieves competitive results against other systems, e.g. in the recent WMT 2017 translation task, it won third place in English-German translation with a single model as shown in Table~\ref{tab:wmt}.  
The system is also competitive in speed as shown in Table~\ref{tab:res}.   Here we compare training and test speed to the publicly available \textit{Nematus} system\footnote{\url{https://github.com/rsennrich/nematus} Comparison with
  OpenNMT/Nematus github revisions {\tt 907824}/{\tt 75c6ab1}} on English-to-German
(EN$\rightarrow$DE) using the WMT2015\footnote{\url{http://statmt.org/wmt15}} dataset. 

We have found that OpenNMT's default setting is useful for experiments, but not optimal for large-scale NMT. This has been a cause of poor reported  performance in other default comparisons by \cite{britz2017massive} and \cite{hieber2017sockeye}. 
We trained models with our best effort to conform to their settings and report our results in Table~\ref{tab:results-updated}, which shows comparable performance with other systems. We suspect that the reported poor performance is due to the fact that our default setting discards sequences of length greater than 50, which is too short for BPE. Moreover, while the reported poor performance was obtained by training with ADAM, we find that training with (the default) SGD with learning rate decay is generally better.

We also compare OpenNMT with the \textit{GNMT} \citep{wu2016google} model in Table~\ref{tab:results-google}. 
\cite{vaswani2017attention} have established a new state-of-the-art with the Transformer model. We have also implemented this in our framework, and compare it with \textit{Tensor2Tensor} (T2T) in Table~\ref{tab:results-transformer}. (These experiments are run on a modified version of WMT 2017, namely News Comm v11 instead of v12, and no Rapid 2016.)

Additionally we have found interest from the community in using
OpenNMT for language geneation tasks like sentence document summarization and
dialogue response generation, among others.  Using
OpenNMT, we were able to replicate the sentence summarization results
of \citet{chopra2016abstractive}, reaching a ROUGE-1 score of 35.51 on
the Gigaword data. We have also trained a model on 14 million
sentences of the OpenSubtitles data set based on the work
\citet{vinyals2015neural}, achieving comparable perplexity. Many other models are at \url{http://opennmt.net/Models-py} and \url{http://opennmt.net/Models}.

\newpage
\small

\bibliographystyle{apalike}
\bibliography{mail.bbl}

\begin{thebibliography}{}

\bibitem[Ameur et~al., 2017]{ameur2017arabic}
Ameur, M. S.~H., Meziane, F., and Guessoum, A. (2017).
\newblock Arabic machine transliteration using an attention-based
  encoder-decoder model.
\newblock {\em Procedia Computer Science}, 117:287--297.

\bibitem[Bahdanau et~al., 2014]{Bahdanau2015}
Bahdanau, D., Cho, K., and Bengio, Y. (2014).
\newblock {Neural Machine Translation By Jointly Learning To Align and
  Translate}.
\newblock In {\em ICLR}, pages 1--15.

\bibitem[Britz et~al., 2017]{britz2017massive}
Britz, D., Goldie, A., Luong, T., and Le, Q. (2017).
\newblock Massive exploration of neural machine translation architectures.
\newblock {\em arXiv preprint arXiv:1703.03906}.

\bibitem[Chan et~al., 2015]{DBLP:journals/corr/ChanJLV15}
Chan, W., Jaitly, N., Le, Q.~V., and Vinyals, O. (2015).
\newblock Listen, attend and spell.
\newblock {\em CoRR}, abs/1508.01211.

\bibitem[Cho et~al., 2014]{Cho2014}
Cho, K., van Merrienboer, B., Gulcehre, C., Bahdanau, D., Bougares, F.,
  Schwenk, H., and Bengio, Y. (2014).
\newblock {L}earning {P}hrase {R}epresentations using {RNN} {E}ncoder-{D}ecoder
  for {S}tatistical {M}achine {T}ranslation.
\newblock In {\em Proc of EMNLP}.

\bibitem[Chopra et~al., 2016]{chopra2016abstractive}
Chopra, S., Auli, M., and Rush, A.~M. (2016).
\newblock Abstractive sentence summarization with attentive recurrent neural
  networks.
\newblock {\em Proceedings of NAACL-HLT16}, pages 93--98.

\bibitem[Chung et~al., 2014]{chung2014empirical}
Chung, J., Gulcehre, C., Cho, K., and Bengio, Y. (2014).
\newblock Empirical evaluation of gated recurrent neural networks on sequence
  modeling.
\newblock {\em arXiv preprint arXiv:1412.3555}.

\bibitem[Crego et~al., 2016]{systran}
Crego, J., Kim, J., and Senellart, J. (2016).
\newblock Systran's pure neural machine translation system.
\newblock {\em arXiv preprint arXiv:1602.06023}.

\bibitem[Dean et~al., 2012]{dean2012large}
Dean, J., Corrado, G., Monga, R., Chen, K., Devin, M., Mao, M., Senior, A.,
  Tucker, P., Yang, K., Le, Q.~V., et~al. (2012).
\newblock Large scale distributed deep networks.
\newblock In {\em Advances in neural information processing systems}, pages
  1223--1231.

\bibitem[Deng et~al., 2017]{deng2017image}
Deng, Y., Kanervisto, A., Ling, J., and Rush, A.~M. (2017).
\newblock Image-to-markup generation with coarse-to-fine attention.
\newblock In {\em International Conference on Machine Learning}, pages
  980--989.

\bibitem[Dyer et~al., 2010]{dyer2010cdec}
Dyer, C., Weese, J., Setiawan, H., Lopez, A., Ture, F., Eidelman, V.,
  Ganitkevitch, J., Blunsom, P., and Resnik, P. (2010).
\newblock cdec: A decoder, alignment, and learning framework for finite-state
  and context-free translation models.
\newblock In {\em Proceedings of the ACL 2010 System Demonstrations}, pages
  7--12. Association for Computational Linguistics.

\bibitem[Gardent et~al., 2017]{gardent2017webnlg}
Gardent, C., Shimorina, A., Narayan, S., and Perez-Beltrachini, L. (2017).
\newblock The webnlg challenge: Generating text from rdf data.
\newblock In {\em Proceedings of the 10th International Conference on Natural
  Language Generation}, pages 124--133.

\bibitem[Gehring et~al., 2017]{gehring2017convolutional}
Gehring, J., Auli, M., Grangier, D., Yarats, D., and Dauphin, Y.~N. (2017).
\newblock Convolutional sequence to sequence learning.
\newblock {\em arXiv preprint arXiv:1705.03122}.

\bibitem[Gu et~al., 2016]{gu2016incorporating}
Gu, J., Lu, Z., Li, H., and Li, V.~O. (2016).
\newblock Incorporating copying mechanism in sequence-to-sequence learning.
\newblock {\em arXiv preprint arXiv:1603.06393}.

\bibitem[Ha et~al., 2017]{ha2017effective}
Ha, T.-L., Niehues, J., and Waibel, A. (2017).
\newblock Effective strategies in zero-shot neural machine translation.
\newblock {\em arXiv preprint arXiv:1711.07893}.

\bibitem[Hieber et~al., 2017]{hieber2017sockeye}
Hieber, F., Domhan, T., Denkowski, M., Vilar, D., Sokolov, A., Clifton, A., and
  Post, M. (2017).
\newblock Sockeye: A toolkit for neural machine translation.
\newblock {\em arXiv preprint arXiv:1712.05690}.

\bibitem[Hochreiter and Schmidhuber, 1997]{hochreiter1997long}
Hochreiter, S. and Schmidhuber, J. (1997).
\newblock Long short-term memory.
\newblock {\em Neural computation}, 9(8):1735--1780.

\bibitem[Kim et~al., 2017]{kim2017structured}
Kim, Y., Denton, C., Hoang, L., and Rush, A.~M. (2017).
\newblock Structured attention networks.
\newblock {\em arXiv preprint arXiv:1702.00887}.

\bibitem[Koehn et~al., 2007]{koehn2007moses}
Koehn, P., Hoang, H., Birch, A., Callison-Burch, C., Federico, M., Bertoldi,
  N., Cowan, B., Shen, W., Moran, C., Zens, R., et~al. (2007).
\newblock Moses: Open source toolkit for statistical machine translation.
\newblock In {\em Proc ACL}, pages 177--180. Association for Computational
  Linguistics.

\bibitem[Levin et~al., 2017]{levin2017toward}
Levin, P., Dhanuka, N., Khalil, T., Kovalev, F., and Khalilov, M. (2017).
\newblock Toward a full-scale neural machine translation in production: the
  booking. com use case.
\newblock {\em arXiv preprint arXiv:1709.05820}.

\bibitem[Ling and Rush, 2017]{ling2017coarse}
Ling, J. and Rush, A. (2017).
\newblock Coarse-to-fine attention models for document summarization.
\newblock In {\em Proceedings of the Workshop on New Frontiers in
  Summarization}, pages 33--42.

\bibitem[Luong et~al., 2015a]{Luong2015}
Luong, M.-T., Pham, H., and Manning, C.~D. (2015a).
\newblock {E}ffective {A}pproaches to {A}ttention-based {N}eural {M}achine
  {T}ranslation.
\newblock In {\em Proc of EMNLP}.

\bibitem[Luong et~al., 2015b]{luong2015b}
Luong, M.-T., Sutskever, I., Le, Q., Vinyals, O., and Zaremba, W. (2015b).
\newblock {A}ddressing the {R}are {W}ord {P}roblem in {N}eural {M}achine
  {T}ranslation.
\newblock In {\em Proc of ACL}.

\bibitem[Ma et~al., 2017]{ma2017osu}
Ma, M., Li, D., Zhao, K., and Huang, L. (2017).
\newblock Osu multimodal machine translation system report.
\newblock {\em arXiv preprint arXiv:1710.02718}.

\bibitem[Martins and Astudillo, 2016]{martins2016softmax}
Martins, A.~F. and Astudillo, R.~F. (2016).
\newblock From softmax to sparsemax: A sparse model of attention and
  multi-label classification.
\newblock {\em arXiv preprint arXiv:1602.02068}.

\bibitem[Sennrich and Haddow, 2016]{sennrich2016linguistic}
Sennrich, R. and Haddow, B. (2016).
\newblock Linguistic input features improve neural machine translation.
\newblock {\em arXiv preprint arXiv:1606.02892}.

\bibitem[Sennrich et~al., 2015]{DBLP:journals/corr/SennrichHB15}
Sennrich, R., Haddow, B., and Birch, A. (2015).
\newblock Neural machine translation of rare words with subword units.
\newblock {\em CoRR}, abs/1508.07909.

\bibitem[Sutskever et~al., 2014]{sutskever14sequence}
Sutskever, I., Vinyals, O., and Le, Q.~V. (2014).
\newblock {Sequence to Sequence Learning with Neural Networks}.
\newblock In {\em NIPS}, page~9.

\bibitem[van Noord and Bos, 2017]{van2017neural}
van Noord, R. and Bos, J. (2017).
\newblock Neural semantic parsing by character-based translation: Experiments
  with abstract meaning representations.
\newblock {\em arXiv preprint arXiv:1705.09980}.

\bibitem[Vaswani et~al., 2017]{vaswani2017attention}
Vaswani, A., Shazeer, N., Parmar, N., Uszkoreit, J., Jones, L., Gomez, A.~N.,
  Kaiser, L., and Polosukhin, I. (2017).
\newblock Attention is all you need.
\newblock {\em arXiv preprint arXiv:1706.03762}.

\bibitem[Vinyals et~al., 2015]{vinyals2015pointer}
Vinyals, O., Fortunato, M., and Jaitly, N. (2015).
\newblock Pointer networks.
\newblock In {\em Advances in Neural Information Processing Systems}, pages
  2692--2700.

\bibitem[Vinyals and Le, 2015]{vinyals2015neural}
Vinyals, O. and Le, Q. (2015).
\newblock A neural conversational model.
\newblock {\em arXiv preprint arXiv:1506.05869}.

\bibitem[Wiseman et~al., 2017]{wiseman2017challenges}
Wiseman, S., Shieber, S.~M., and Rush, A.~M. (2017).
\newblock Challenges in data-to-document generation.
\newblock {\em arXiv preprint arXiv:1707.08052}.

\bibitem[Wu et~al., 2016]{wu2016google}
Wu, Y., Schuster, M., Chen, Z., Le, Q.~V., Norouzi, M., Macherey, W., Krikun,
  M., Cao, Y., Gao, Q., Macherey, K., et~al. (2016).
\newblock Google's neural machine translation system: Bridging the gap between
  human and machine translation.
\newblock {\em arXiv preprint arXiv:1609.08144}.

\bibitem[Xu et~al., 2015]{xu2015show}
Xu, K., Ba, J., Kiros, R., Cho, K., Courville, A., Salakhudinov, R., Zemel, R.,
  and Bengio, Y. (2015).
\newblock Show, attend and tell: Neural image caption generation with visual
  attention.
\newblock In {\em International Conference on Machine Learning}, pages
  2048--2057.

\end{thebibliography}

\end{document}